\documentclass[10pt,twocolumn,letterpaper]{article}

\usepackage{iccv}
\usepackage{times}
\usepackage{epsfig}
\usepackage{graphicx}
\usepackage{amsmath}
\usepackage{amssymb}
\usepackage{arydshln}

\usepackage{booktabs}
\usepackage{amsthm}
\usepackage{algorithm}
\usepackage{algorithmic}
\usepackage{caption}
\usepackage{subcaption}
\usepackage{booktabs}
\usepackage{multirow}
\usepackage{color}
\newtheorem{lemma}{Lemma}[section]
\usepackage[accsupp]{axessibility}  

\usepackage[breaklinks=true,bookmarks=false]{hyperref}

\iccvfinalcopy 

\def\iccvPaperID{9163} 

\ificcvfinal\fi

\begin{document}


\title{Estimator Meets Equilibrium Perspective:\\ A Rectified Straight Through Estimator for Binary Neural Networks Training}

\author{Xiao-Ming Wu$^{1}$, Dian Zheng$^{1}$, Zuhao Liu$^{1}$, Wei-Shi Zheng$^{1,2,3,4}$\thanks{~denotes the corresponding author.} \\
\normalsize $^1$School of Computer Science and Engineering, Sun Yat-sen University, China, 
\normalsize $^2$Pengcheng Lab, China, \\ 
\normalsize $^3$Guangdong Province Key Laboratory of Information Security Technology, China, \\ 
\normalsize $^4$Key Laboratory of Machine Intelligence and Advanced Computing, Ministry of Education, China\\
{\tt\small \{wuxm65, zhengd35, liuzh327\}@mail2.sysu.edu.cn}, {\tt\small wszheng@ieee.org}}

\maketitle
\ificcvfinal\thispagestyle{empty}\fi

\begin{abstract}
   Binarization of neural networks is a dominant paradigm in neural networks compression. The pioneering work BinaryConnect uses Straight Through Estimator (STE) to mimic the gradients of the sign function, but it also causes the crucial inconsistency problem. Most of the previous methods design different estimators instead of STE to mitigate it. However, they ignore the fact that when reducing the estimating error, the gradient stability will decrease concomitantly. These highly divergent gradients will harm the model training and increase the risk of gradient vanishing and gradient exploding. To fully take the gradient stability into consideration, we present a new perspective to the BNNs training, regarding it as the equilibrium between the estimating error and the gradient stability. In this view, we firstly design two indicators to quantitatively demonstrate the equilibrium phenomenon. In addition, in order to balance the estimating error and the gradient stability well, we revise the original straight through estimator and propose a power function based estimator, \textbf{Re}ctified \textbf{S}traight \textbf{T}hrough \textbf{E}stimator (\textbf{ReSTE} for short). Comparing to other estimators, ReSTE is rational and capable of flexibly balancing the estimating error with the gradient stability. Extensive experiments on CIFAR-10 and ImageNet datasets show that ReSTE has excellent performance and surpasses the state-of-the-art methods without any auxiliary modules or losses. 
\end{abstract}

\section{Introduction}
\label{sec:introduction}

\begin{figure}[t]
\centering
\includegraphics[width=1\linewidth]{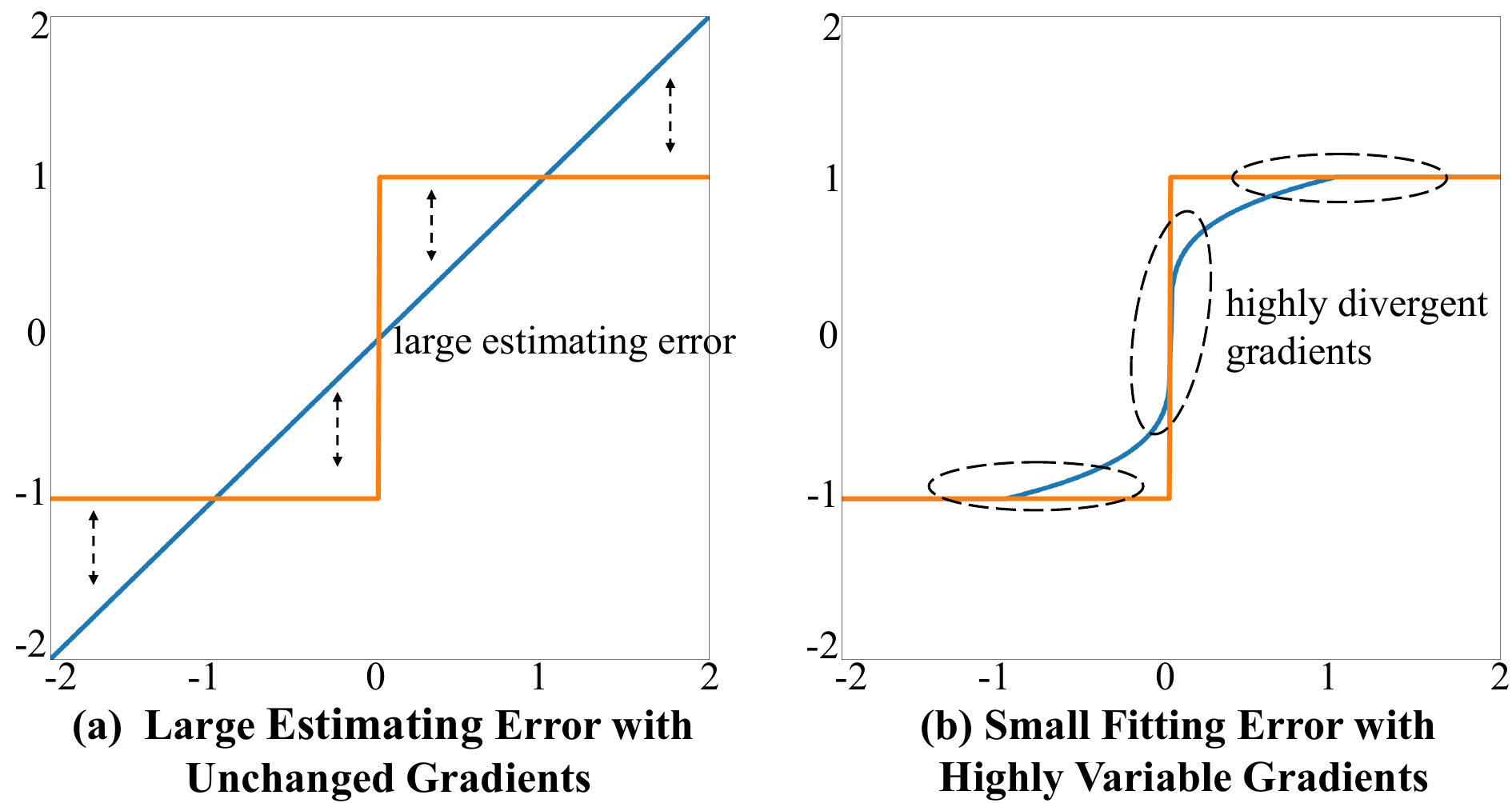}
\caption{The intuitive illustrations of the equilibrium perspective of BNNs training, i.e., the equilibrium between the estimating error and the gradient stability. When reducing the estimating error, the gradients will become highly divergent, which harms the model training and increases the risk of gradient vanishing and gradient exploding. Blue, orange lines represent the estimators and sign function respectively.}
\label{fig:tradeoff}
\end{figure}

Deep neural networks have revolutionary development in recent years since its admirable ability to learn discriminate features~\cite{lecun2015deep, he2016deep, girshick2014rich, long2015fully, redmon2016you}. But they tend to require massive computational cost and memory cost, which is unsuitable to deploy at some resource-limited devices. To this end, many network compression methods have been proposed~\cite{howard2017mobilenets, han2015deep, hinton2015distilling}, such as pruning~\cite{luo2017thinet,zhuang2018discrimination,zhu2021vision,ding2021resrep}, tiny model design~\cite{zhang2018shufflenet,howard2017mobilenets,tan2019efficientnet,iandola2016squeezenet}, distillation~\cite{park2019relational, tian2019contrastive, zhao2022decoupled} and tensor decomposition\cite{rabanser2017introduction}. Among them, network quantization~\cite{courbariaux2015binaryconnect, rastegari2016xnor,  leng2018extremely, helwegen2019latent, ajanthan2021mirror} is a kind of excellent method with high compression ratio and little performance degradation. Binary Neural Networks (BNNs)~\cite{courbariaux2015binaryconnect, courbariaux2016binarized, hubara2017quantized}, an extreme case of network quantization which aims to quantize 32-bit inputs into 1-bit, attract great research enthusiasm in recent years due to its extremely high compression ratio and great performance in neural networks compression.

In BNNs research, the pioneering work BinaryConnect~\cite{courbariaux2015binaryconnect} proposes to apply sign function to binary the full-precision inputs in forward process, and use the straight through estimator (STE) to mimic the gradients of the sign function when backpropagation, which achieves great performance. However, the difference between the forward and the backward processes causes the crucial inconsistency problem in BNNs training. To reduce the degree of inconsistency, many previous works design different estimators instead of STE, attempting to narrow the estimating error. Nevertheless, they neglect the fact that when reducing the estimating error, the gradient stability will decrease concomitantly. This will make the gradients highly divergent, harming the model training and increasing the risk of gradient vanishing and gradient exploding.

To fully take the gradient stability into consideration, we present a new perspective to the BNNs training, regarding it as the equilibrium between the estimating error and the gradient stability, as shown in Fig. \ref{fig:tradeoff}. In this view, we firstly design two indicators to measure the degree of the equilibrium between estimating error and the gradient instability. With these indicators, we can quantitatively demonstrate the equilibrium phenomenon. In addition, to balance the estimating error with the gradient stability well, we revise the original straight through estimator (STE) and propose a power function based estimator, \textbf{Re}ctified \textbf{S}traight \textbf{T}hrough \textbf{E}stimator, \textbf{ReSTE} for short. The insight is from the fact that STE is a special case of the power function. With this design, ReSTE is always rational, i.e., having less estimating error than STE, and capable of flexibly balancing the estimating error and the gradient stability, which are the two main advantages of ReSTE comparing to other estimators.

 Sufficient experiments on CIFAR-10~\cite{krizhevsky2009learning} and large-scale ImageNet ILSVRC-2012~\cite{deng2009ImageNet} datasets show that our method has good performance and surpasses the state-of-the-art methods without any auxiliaries, e.g., additional modules or losses. Moreover, by two carefully-designed indicators, we demonstrate the equilibrium phenomenon and further show that ReSTE can flexibly balance the estimating error and the gradient stability. Our source code is available at \url{https://github.com/DravenALG/ReSTE}, hoping to help the development of the BNNs community.

\section{Revisiting Binary Neural Networks}
\label{sec:related_works}
Binary Neural Networks (BNNs)~\cite{courbariaux2015binaryconnect, courbariaux2016binarized, rastegari2016xnor, cai2017deep} aim to binarize full-precision inputs, weights or features (also called activations in BNNs literature) in each layers into 1-bits, which is an extreme case of network quantization. Essentially, the optimization of BNNs is a constraint optimization problem. Naively using brute-force algorithms to solve this problem is intractable due to the huge combinatorial probabilities when the dimensions of input are large.

The exploration of tractable solutions to binary neural networks training can be traced back to many pioneering works~\cite{hwang2014fixed, cheng2015training, soudry2014expectation}. Among them, BinaryConnect~\cite{courbariaux2015binaryconnect} forms the main optimization paradigm in this domain due to its great performance. BinaryConnect connects a sign function between the full-precision inputs and the following calculation modules in forward process. Since the gradients of the sign function are zero almost everywhere, BinaryConnect uses an identity function to substitute for the sign function when calculating the gradients in backward process, which is also known as straight through estimator (STE)~\cite{hinton2012neural,bengio2013estimating}. For convenience, we respectively donate $\mathbf{z}$ and $\mathbf{z}_b$ as the full-precision inputs and the binarized outputs. The forward and backward processes of the binary procedure in BinaryConnect are as follows:
\begin{equation}\label{STE_identity}
\begin{aligned}
\text{Forward:} \mathbf{z_b} = \mathbf{sign(z) },
\\
\text{Backward:} \frac{\partial \mathcal{L} }{\partial \mathbf{z}} = \frac{\partial \mathcal{L} }{\partial \mathbf{z}_b},
\end{aligned} 
\end{equation}
where $\mathcal{L} $ is the loss function and $\mathbf{sign}$ represents the element-wise sign function. It means that the gradients with respect to the full-precision inputs straightly equals to the gradients of the binarized outputs, which is also the origin of the name straight through estimator.

To improve the performance of binary neural networks, many different improvement strategies have been proposed. Some works try to modify the model architectures of the backbone, which heightens the expressive ability of the binary neural networks~\cite{liu2018bi, liu2020reactnet}. In spite of the performance improvement, these works revise the architectures of the backbone, which is not universal to all architectures and adds additional computational and memory cost in inference. In addition, some other works focus on improving the forward process with some additional assistance, e.g., modules~\cite{zhang2018lq, lin2020rotated, xu2021learning, tu2022adabin}, losses~\cite{bai2018proxquant, gu2019projection, gu2019bayesian, liu2021rectified, shang2022lipschitz, xu2022recurrent} and even distillation~\cite{tian2019contrastive}.  This type of methods significantly  increase parameters and the computational cost when training.

Besides, many works mainly focus on the essential and vital component of binary neural networks training, i.e, the estimator to mimic the gradients of the sign function.
BNN+~\cite{darabi2018bnn+} designs a SignSwish function, Bi-Real-Net~\cite{liu2018bi} models a piece-wise polynomial function, DSQ~\cite{gong2019differentiable} proposes a tanh-based function, RQ~\cite{zhang2023root} proposes a root based function similar to our ReSTE but more complex and not focuses on balancing the equilibrium, IR-Net~\cite{qin2020forward} gives the EDE function and FDA~\cite{xu2021learning} applies Fourier series to simulate the gradients. Although they have achieved excellent performance, they ignore the fact that when reducing the estimating error, the gradient stability will decrease concomitantly, which means that the gradient will become highly divergent, harming the model training and increasing the risk of gradient vanishing and gradient exploding. To fully consider the gradient stability in BNNs training, we present a new perspective, viewing it as the equilibrium between the estimating error and the gradient stability. From this perspective, we revise the original STE and propose a power function based estimator, Rectified Straight Through Estimator (ReSTE for short). Comparing to the estimators above, ReSTE is rational, i.e., having less or equal estimating error than STE  and capable of flexibly balancing the estimating error and gradient stability. Sufficient experiments show that our method surpasses the state-of-the-art methods without any auxiliaries, e.g., additional modules or losses.

\begin{figure}[t]
\centering
\includegraphics[width=0.85\linewidth]{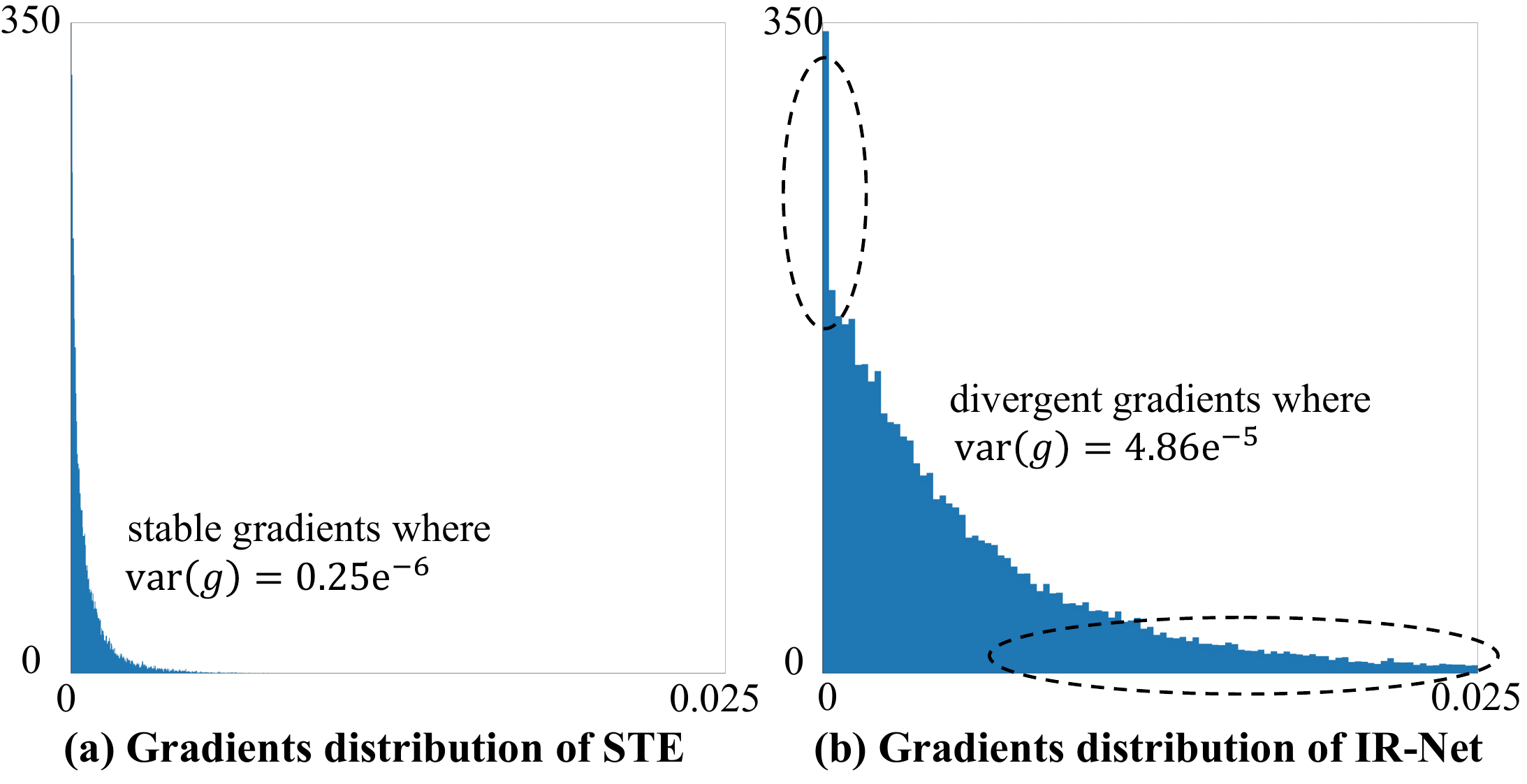}
\caption{Illustrations of the gradient distributions of STE (left) and IR-Net (right). X-axes represent the values of the gradients, y-axes are the frequency.}
\label{fig:STE_IR}
\end{figure}

\section{Estimator Meets Equilibrium Perspective}
\label{sec:approach}
\subsection{Equilibrium Perspective}
\label{sec:equilibrium}
The \textbf{inconsistency problem} is inevitable but crucial in BNNs training since we use estimators to mimic the gradients of sign function in backpropogation. To mitigate the degree of the inconsistency, lots of follow-up works design different estimators instead of STE, aiming to reduce the estimating error. Although they improve the performance of BNNs, they only care about reducing the estimating error and ignore the concomitant gradient instability. The gradients will become highly divergent, which increases the risk of gradients vanishing and gradients exploding, as shown in Fig. \ref{fig:tradeoff}. For more persuasive, we visualize the gradient distributions of STE~\cite{rastegari2016xnor} and the influential work IR-Net~\cite{qin2020forward} in Fig. \ref{fig:STE_IR}. Although IR-Net attempts to reduce the estimating error by approximating the sign function as it claims, it suffers from the problem of highly divergent gradients, which will harm the model training.

To fully take the gradients stability into consideration, we present a new perspective, considering BNNs training as the equilibrium between the estimating error and the gradient stability. For clear description, we first give the definition of the estimating error and the gradient stability. We define that the \textbf{estimating error} is the difference between the sign function and the estimator, which reflects how close between the estimator and sign function. We define the \textbf{gradient stability} as the divergence of the gradients of all parameters in each iteration. The insight is that when we use estimator to close to sign function, the gradients of all parameters in one iteration are inevitably divergent, which is intuitively shown in Fig. \ref{fig:tradeoff}. This may lead to a wrong updating direction and harm the model training.

With the definitions, we now formally discuss our \textbf{equilibrium perspective}. Since the BNNs training is the equilibrium between the estimating error and the gradient stability, we should not reduce estimating error without limits. Instead, we should design an estimator which can easily adjust the degree of equilibrium to obtain better performance.

\subsection{Indicators of Estimating Error and Gradient Instability}
To quantitatively and clearly demonstrate the equilibrium phenomenon, we firstly design two indicators to quantitatively analyze the degree of the estimating error and the gradient instability.

Since the estimating error is the difference between the sign function and the estimator, we stipulate that the estimating error can be evaluated by the distance between the results through the element-wise sign function and the results through the estimator in each iteration. We define $\mathbf{f(\cdot)}$ as the estimator and $D$ as the distance metric. The degree of estimating error can be formally described as: 
\begin{equation}
\begin{aligned}
e = D(\mathbf{sign(z)} ,\mathbf{f(z)} ),
\end{aligned} 
\end{equation}
where $D(\cdot)$ is L2-norm in our method. We call $e$ as the \textbf{estimating error indicator}.

In addition, to measure the degree of the gradient stability, we design a \textbf{gradient instability indicator}. Since the gradient stability is the divergence of the gradients of all parameters in each iteration, we use the variance of gradients of all the parameters in each iteration to evaluate it. We design the indicator as follows:
\begin{equation}
\begin{aligned}
s = \mathrm{var}(\mathbf{|g|})),
\end{aligned}
\end{equation}
where $\mathbf{g}$ donates the gradients, $|\cdot|$ is the element-wise absolute operation and $\mathrm{var(\cdot)}$ stands for the variance operator. Here we use absolute operation since we only care about the gradients magnitude (the updating directions are not relevant to the gradient stability). Note that $s$ is the gradient instability indicator that the magnitude of $s$ reflects the degree of the instability.

\subsection{Rectified Straight Through Estimator}
\label{sec:ReSTE}
To balance the estimating error and the gradient stability, we should design an estimator that can easily adjust the degree of equilibrium well. Before that, we firstly claim that sign function and STE are two extremes in gradient stability. The sign function has zero gradients almost everywhere and has infinite gradients at the origin of the coordinate, whose gradients are completely vanishing or exploding. Therefore sign function has the highest gradient instability. In contract, STE uses linear function to estimate the gradients of sign function, which not at all changes the gradients backward in the estimating process. So STE is with the lowest instability. We want to design an estimator close to sign function to get less estimating error, but not too much unstable to train. Based on this intuition, we design two properties that an estimator should satisfy:
\textbf{1) Rational property:} It should always have less or equal estimating error than straight through estimator (the identity function), which can be formally described as $D(\mathbf{sign(z)}  ,\mathbf{f(z)} )-D(\mathbf{sign(z)}  ,\mathbf{z} )\le 0$. The rational property is rational since the fact that if an estimator has more estimating error than STE in some ranges, directly applying STE to mimic the gradients in these ranges is more reasonable, which is more stable and has less estimating error.
\textbf{2) Flexible property:} It should be capable of flexibly adjusting the degree of the estimating error and the gradient stability to adjust the degree of equilibrium. The flexible property consists of two aspects. First, the estimator can change from STE to sign function. Second, the changing should be gradually. Gradually changing means that each point can move a small step closer to sign function when we adjust the estimator, which is the key to find a suitable degree of the equilibrium.

\begin{figure}[t]
\centering
\includegraphics[width=0.85\linewidth]{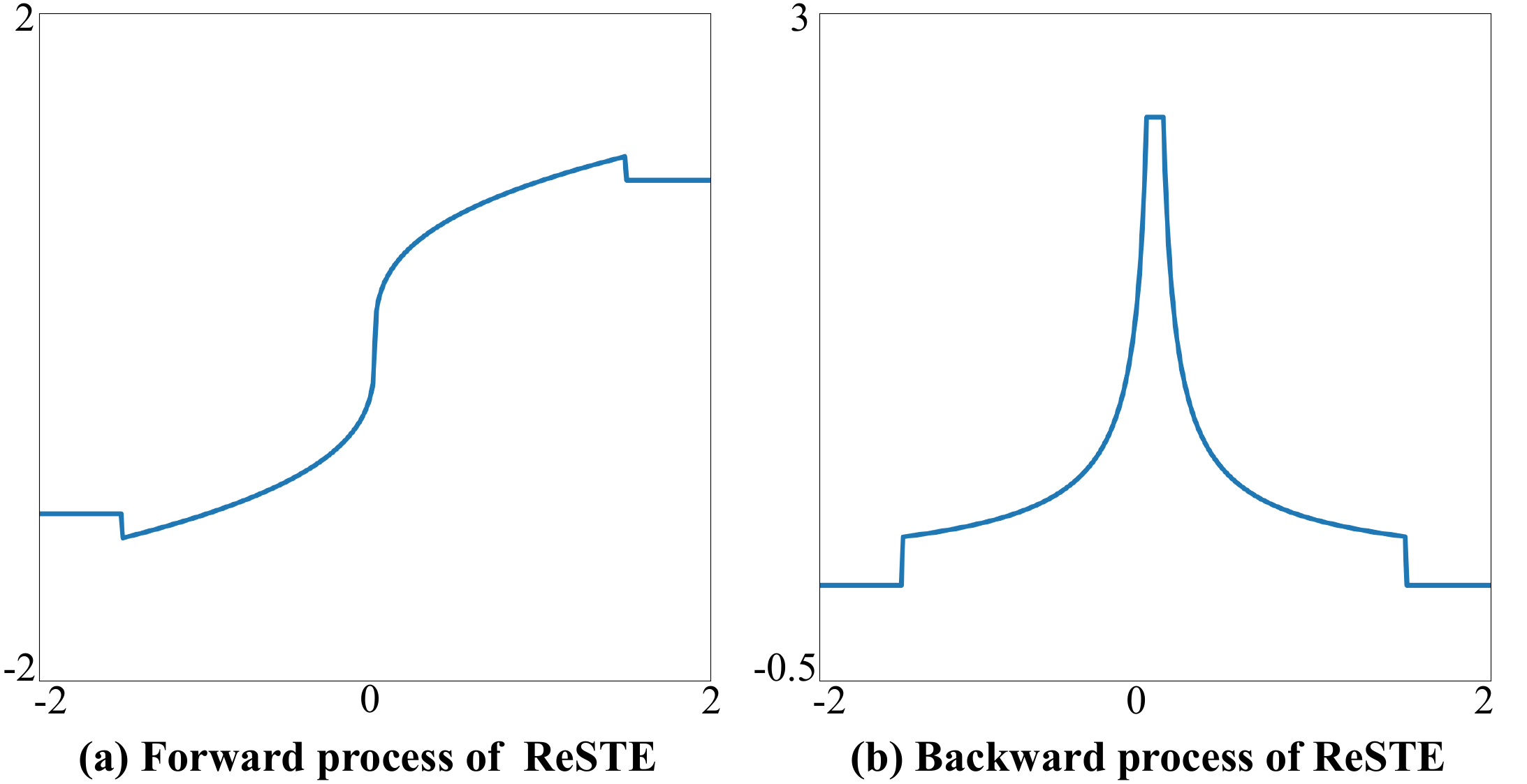}
\caption{Illustrations of the forward (left) and backward (right) processes of ReSTE. }
\label{fig:ReSTE}
\end{figure}

To achieve these goals, we revise the STE and propose a power function based estimator, \textbf{Re}ctified \textbf{S}traight \textbf{T}hrough \textbf{E}stimator, \textbf{ReSTE} for short. The inspiration of ReSTE is that the STE strategy (identity function) is a special case of the power function, when the power is one for specific. When the power function is close to STE, the gradient is stable, but the estimating error is large. When the power increases, the power function will close to sign function and have less estimating error, while increasing the instability of the gradients. In a word, power function can easily change from STE to sign function, demonstrating its ability of adjusting the degree of equilibrium.

Under such observation, we propose to use power function as the estimator in backward process to balance the estimating error and the gradient stability. Our ReSTE function has the following form:
\begin{equation}
\begin{aligned}
&\mathbf{f(z)} =\mathbf{sign(z)}|\mathbf{z}|^{\frac1o},
\\&s.t. \quad o \geq 1,
\end{aligned}
\end{equation}
where $o$ are hyper-parameters controlling the power, which is also the degree of the equilibrium. In detail, $o$ decides the ratified degree of ReSTE to balance the estimating error and gradient stability. Note that when $o=1$, the ReSTE function is the basic STE. With o increasing, the ReSTE function closes to sign function, which has less estimating error gradually. With simple derivation, the gradients of the ReSTE function is:
\begin{equation}\label{Backfard_ReSTE}
\begin{aligned}
\mathbf{f'(z)} & = \frac{1}{o}|\mathbf{z}|^{\frac{1-o}{o}}.
\end{aligned}
\end{equation}

Comparing to other estimators, ReSTE satisfies the properties proposed above, i.e., rational and capable of flexibly balancing the estimating error and the gradient stability, which are the two main advantages of our method. To prove that, we firstly give the following lemma.
\begin{lemma}\label{lemma:power_function}
If $o_1 \ge o_2$, $D(\mathbf{sign(z)}  ,\mathbf{f}(\mathbf{z}, o_1) )-D(\mathbf{sign(z)}  ,\mathbf{f}(\mathbf{z}, o_2) )\le 0$ holds. The \textit{proof} is as follows:
\end{lemma}
\begin{equation}
\begin{aligned}
&D(\mathbf{sign(z)},\mathbf{f}(\mathbf{z}, o_1) )  \\
&= \sqrt{\sum_{i = 1}^{d}{(\mathrm{sign} (z_i)-\mathrm{f}(z_i, o_1) )^2}} \\
&= \sqrt{\sum_{i = 1}^{d}{(\mathrm{sign} (z_i)-\mathrm{sign} (z_i)|z_i|^{\frac{1}{o_1} } )^2}} \\
&= \sqrt{\sum_{i = 1}^{d}{|1-|z_i|^{\frac{1}{o_1} } |^2}},
\end{aligned}
\end{equation}
where $|\cdot|$ is the absolute operation. Since $o_1 \ge o_2$, with the nature of the power function, we can achieve that when $|z_i|\le1$, $|1-|z_i|^{\frac{1}{o_1}}|=1-|z_i|^{\frac{1}{o_1}}\le1-|z_i|^{\frac{1}{o_2}}=|1-|z_i|^{\frac{1}{o_2}}|$, and when $|z_i|\ge1$, $|1-|z_i|^{\frac{1}{o_1}}|=|z_i|^{\frac{1}{o_1}}-1\le|z_i|^{\frac{1}{o_2}}-1=|1-|z_i|^{\frac{1}{o_2}}|$. Thus, $|1-|z_i|^{\frac{1}{o_1}}|\le|1-|z_i|^{\frac{1}{o_2}}|$ always holds. Then, we can write:
\begin{equation}
\begin{aligned}
D(\mathbf{sign(z)},\mathbf{f(z)} ) &= \sqrt{\sum_{i = 1}^{d}{|1-|z_i|^{\frac{1}{o_1} } |^2}} \\
& \le \sqrt{\sum_{i = 1}^{d}{|1-|z_i |^{\frac{1}{o_2} }|}^2}\\
& = D(\mathbf{sign(z)},\mathbf{f}(\mathbf{z}, o_2) ).
\end{aligned}
\end{equation}

Under the lemma, we give the proof of the two properties. As for the rational property, since STE equals to $\mathbf{f}(\mathbf{z}, 1)$ and ReSTE has the condition $o \ge 1$, we can easily get that $D(\mathbf{sign(z)}  ,\mathbf{f(z)} )-D(\mathbf{sign(z)}  ,\mathbf{z} )\le 0$ always holds by lemma \ref{lemma:power_function}. About the flexible property, we know that STE equals to $\mathbf{f}(\mathbf{z}, 1)$ and when $o\to \infty$, $\mathbf{f(z)} \to \mathbf{sign(z)}$, thus ReSTE can change from STE to sign function. Moreover, from the proof of lemma \ref{lemma:power_function} we can observe that if $o_1 \ge o_2$, $|1-|z_i|^{\frac{1}{o_1}}|\le|1-|z_i|^{\frac{1}{o_2}}|$ always holds for any $z_i$, thus $|\mathrm{sign} (z_i)-\mathrm{f}(z_i, o_1)|\le |\mathrm{sign} (z_i)-\mathrm{f}(z_i, o_2)| $ always holds for any $z_i$. So the changing of ReSTE is gradually, where any $z_i$ moves a small step closer to sign function when increasing $o$. Therefore, ReSTE satisfies the flexible property. The rational and flexible properties are designed based on the equilibrium perspective and form the main advantages between ReSTE and other estimators in previous methods.

In addition, for more stable gradients, we use some gradients truncation tricks to our estimator. First, we clip the gradients where the corresponding full-precision inputs with the absolute value larger than a threshold $t$ to zero, which considers the saturation in BNNs training~\cite{courbariaux2016binarized}. Next, since the gradients of ReSTE may be large when the input is sufficiently small, we set a threshold $m$ and the gradients within the threshold $(0, m), (-m, 0)$ use numerical method $(f(m)-f(0))/m, (f(0)-f(-m))/m$ to simulate. 

For clear illustration, we demonstrate the forward and backward processes of ReSTE in Fig. \ref{fig:ReSTE}.

\begin{table}[t]
\centering
\scriptsize
\begin{tabular}{clccc}\toprule
Backbone & Method & W/A & Auxiliary & Acc(\%)\\ \midrule
\multirow{6}{*}{ResNet-18} & FP & 32/32 & - & 94.84\\
& RAD~\cite{ding2019regularizing} & 1/1 & Loss & 90.50\\
& IR-Net~\cite{qin2020forward} & 1/1 & Module & 91.50\\
& LCR-BNN~\cite{shang2022lipschitz} & 1/1 & Loss & 91.80\\
& RBNN~\cite{lin2020rotated} & 1/1 & Module & 92.20\\
& ReSTE (ours) & 1/1 & - & \textbf{92.63}\\\midrule
\multirow{18}{*}{ResNet-20} & FP & 32/32 & - & 91.70\\
& DSQ~\cite{gong2019differentiable} & 1/1 & - & 84.11\\
& DoReFa-Net~\cite{zhou2016dorefa} & 1/1 & - & 84.44\\
& IR-Net~\cite{qin2020forward} & 1/1 & Module & 85.40\\
& LCR-BNN~\cite{shang2022lipschitz} & 1/1 & Loss & 86.00\\
& FDA & 1/1 & Module & 86.20\\
& RBNN~\cite{lin2020rotated} & 1/1 & Module & 86.50\\
& ReSTE (ours) & 1/1 & - & \textbf{86.75}\\ \cdashline{2-5}
& IR-Net *~\cite{qin2020forward} & 1/1 & Module & 86.50\\
& LCR-BNN *~\cite{shang2022lipschitz} & 1/1 & Loss &  87.20\\
& RBNN *~\cite{lin2020rotated} & 1/1 & Module &  87.50\\
& ReSTE * (ours) & 1/1 & - &  \textbf{87.92}\\\cmidrule(lr){2-5}
& FP & 32/32 & - &  91.70\\
& DoReFa-Net~\cite{zhou2016dorefa} & 1/32 & - & 90.00\\
& LQ-Net~\cite{zhang2018lq} & 1/32 & - & 90.10\\
& DSQ~\cite{gong2019differentiable} & 1/32 & - & 90.20\\
& IR-Net~\cite{qin2020forward} & 1/32 & Module &  90.80\\
& LCR-BNN~\cite{shang2022lipschitz} & 1/32 & Loss &  91.20\\
& ReSTE (ours) & 1/32 & - & \textbf{91.32}\\\midrule
\multirow{8}{*}{VGG-small} & FP & 32/32 & - & 93.33\\
& LBA~\cite{hou2016loss} & 1/1 & - &  87.70\\
& Xnor-Net~\cite{rastegari2016xnor} & 1/1 & - &  89.80\\
& BNN~\cite{courbariaux2016binarized} & 1/1 & - & 89.90\\
& RAD~\cite{ding2019regularizing} & 1/1 & Loss &  90.00\\
& IR-Net~\cite{qin2020forward} & 1/1 & Module & 90.40\\
& RBNN~\cite{lin2020rotated} & 1/1 & Module &  91.30\\
& ReSTE (ours) & 1/1 & - &  \textbf{92.55}\\
\bottomrule
\end{tabular}
\caption{Performance comparison with SOTA methods in CIFAR-10 dataset. Auxiliary refers to whether some additional assistance is used (module or loss). FP is the full-precision version of the backbone. * donates the method with Bi-Real structure. W/A is the bit width of weights or activations. Best results are shown in black bold font.}
\label{CIFAR-10_performance}
\end{table}

\begin{table}[t]
\centering
\scriptsize
\resizebox{\linewidth}{!}{
\begin{tabular}{clcccc}\toprule
Backbone & Method & W/A & Auxiliary & Top-1(\%) & Top-5(\%)\\ \midrule
\multirow{22}{*}{ResNet-18} & FP & 32/32 & - & 69.60 & 89.20\\
& ABC-Net~\cite{lin2017towards} & 1/1 & - & 42.70 & 67.60\\
& Xnor-Net~\cite{rastegari2016xnor} & 1/1 & - & 51.20 & 73.20\\
& BNN+~\cite{bulat2019xnor} & 1/1 & Loss & 53.00 & 72.60\\
& DoReFa-Net~\cite{zhou2016dorefa} & 1/2 & - & 53.40 & -\\
& Bi-Real~\cite{liu2018bi} & 1/1 & - & 56.40 & 79.50\\
& Xnor-Net++~\cite{bulat2019xnor} & 1/1 & -  &  57.10 & 79.90\\
& IR-Net~\cite{qin2020forward} & 1/1 & Module &  58.10 & 80.00\\
& LCR-BNN~\cite{shang2022lipschitz} & 1/1 & Loss &  59.60 & 81.60\\
& RBNN~\cite{lin2020rotated} & 1/1 & Module &  59.90 & 81.90\\
& FDA~\cite{xu2021learning} & 1/1 & Module &  60.20 & 82.30\\
& ReSTE (ours) & 1/1 & - &  \textbf{60.88} & \textbf{82.59}\\\cmidrule(lr){2-6}
& FP & 32/32 & - &  69.60 & 89.20\\
& SQ-BWN~\cite{dong2017learning} & 1/32 & - &  58.40 & 81.60\\
& BWN~\cite{rastegari2016xnor} & 1/32 & - &  60.80 & 83.00\\
& HWGQ~\cite{cai2017deep} & 1/32 & - &  61.30 & 83.20\\
& TWN~\cite{alemdar2017ternary} & 2/32 & - &  61.80 & 84.20\\
& SQ-TWN~\cite{dong2017learning} & 2/32 & - &  63.80 & 85.70\\
& BWHN~\cite{hu2018hashing} & 1/32 & - &  64.30 & 85.90\\
& IR-Net~\cite{qin2020forward} & 1/32 & Module &  66.50 & 86.80\\
& LCR-BNN~\cite{shang2022lipschitz} & 1/32 & Loss &  66.90 & 86.40\\
& ReSTE (ours) & 1/32 & - &  \textbf{67.40} & \textbf{87.20}\\
\midrule
\multirow{10}{*}{ResNet-34} & FP & 32/32 & - &  73.30 & 91.30\\
& ABC-Net~\cite{lin2017towards} & 1/1 & - &  52.40 & 76.50\\
& Bi-Real~\cite{liu2018bi} & 1/1 & - &  62.20 & 83.90\\
& IR-Net~\cite{qin2020forward} & 1/1 & Module &  62.90 & 84.10\\
& RBNN~\cite{lin2020rotated} & 1/1 & Module &  63.10 & 84.40\\
& LCR-BNN~\cite{shang2022lipschitz} & 1/1 & Loss &  63.50 & 84.60\\
& ReSTE(ours) & 1/1 & - & \textbf{ 65.05} & \textbf{85.78} \\
\cmidrule(lr){2-6}
& FP & 32/32 & - & 73.30 & 91.30\\
& IR-Net~\cite{qin2020forward} & 1/32 & Module & 70.40 & \textbf{89.50}\\
& ReSTE(ours) & 1/32 & - &  \textbf{70.74} & \textbf{89.50}\\
\bottomrule
\end{tabular}}
\caption{Performance comparison with SOTA methods in ImageNet dataset. Auxiliary refers to whether some additional assistance is used  (module or loss). FP is the full-precision version of the backbone.  W/A is the bit width of weights or activations. Best results are in black bold font.}
\label{ImageNet_performance}
\end{table}

\begin{table*}[t]
\centering
\scriptsize
\begin{tabular}{ccccccc}\toprule
Estimators & Formula & Type & Rational & Flexible  & Acc(\%)\\ \midrule
DSQ~\cite{gong2019differentiable} & $\mathbf{f(z)}=l+\Delta\left(\mathbf{i} + (s \mathbf{tanh} \left(k\left(\mathbf{z}-\mathbf{m}\right)\right)+1)/2\right)$ & Tanh-alike & Not rational & Little flexible &   84.11\\\midrule

STE~\cite{zhou2016dorefa} & $\mathbf{f(z)} =\mathbf{z}$ & Identity function & Rational & Not flexible &   84.44\\\midrule

EDE~\cite{qin2020forward} & $\mathbf{f(z)} =k \mathbf{tanh} (t \mathbf{z})$ & Tanh-alike & Not rational & Little flexible &  85.20\\\midrule

FDA $\dagger$ ~\cite{xu2021learning} & $\mathbf{f(z)} =\frac{4}{\pi} \sum_{i=0}^{k} \mathbf{sin} ((2 i+1) \omega \mathbf{z}) / (2i+1)$& Fourier series & Not rational & Little flexible &  85.80\\\midrule

RBNN $\dagger$ ~\cite{lin2020rotated} & $\mathbf{f(z)} = k \cdot\left(-\mathbf{sign(z)} \frac{t^{2} \mathbf{z}^{2}}{2}+\sqrt{2} t \mathbf{z}\right)$ & Polynomial function & Not rational  & Little flexible & 85.87 \\\midrule

ReSTE (ours) & $\mathbf{f(z)} =\mathbf{sign(z)}|\mathbf{z}|^{\frac1o}$ & Power function & Rational & Flexible &  \textbf{86.75}\\

\bottomrule
\end{tabular}
\caption{Results of the estimators comparison. $\dagger$ means we only use the estimators for fair comparison (without some additional modules, the overall comparison can be found in Sec. \ref{sec:performance}). "Rational" means that the estimator satisfies the rational property proposed in Sec. \ref{sec:ReSTE} while "Not rational" indicates dissatisfaction. "Flexible" means that the estimator satisfies the flexible property proposed in Sec. \ref{sec:ReSTE} while "Not flexible" and "Little flexible" means dissatisfaction. "Not flexible" implies that the estimator can not reduce the estimator. "Little flexible" indicates that the estimator can reduce the estimating error in some kind but not fully satisfy the flexible property. The best results are shown in black bold font.}
\label{ablation_table}
\end{table*}

\subsection{Overall Binary Method}
We summarize the overall Binary procedure of our method. As for the forward process of binarization, we employ DoReFa-Net~\cite{zhou2016dorefa} as most of the previous methods do\cite{qin2020forward, lin2020rotated, xu2021learning, shang2022lipschitz}, which uses sign function to binarize the inputs and endows a layer-level scalar $\beta = \left \|  \mathbf{z} \right \|_{l 1} / {n}$ ($n$ is the dimensions of $\mathbf{z}$) for binarization to enhance the representative ability. In backpropagation, we apply ReSTE as the estimator to simulate the gradients of the sign function. About the hyper-parameter $o$ to adjust the degree of equilibrium, we use the progressive adjusting strategy, which is proposed in \cite{qin2020forward} and widely used in recent works\cite{lin2020rotated, xu2021learning}. We change $o$ from 1 to $o_{\text{end}}$ when training, which we use $o_{\text{end}}=3$ in our experiments. Comparing to the fixed strategy, the progressive adjusting strategy ensures sufficient updating at the beginning and accurate gradients at the end of the training. Experiments about the design for the tuning strategies of $o$ are shown in supplementary materials.

In BNNs literature, there have two types of options to binary a neural network. The first type is that only the weights are binarize and the second type is weights and activations are both to be binarized, which significantly improves the inference speed via XNOR and Bitcount operations~\cite{courbariaux2016binarized, qin2020forward}. After binarization, the model size decreases 32x comparing to the original full-precision model and the inference process is accelerated.

\section{Experiments}
\label{sec:experiments}
\subsection{Datasets and Settings}
\textbf{Datasets.} In this work, we choose two datasets, i.e. CIFAR-10~\cite{krizhevsky2009learning} and ImageNet ILSVRC-2012~\cite{deng2009ImageNet}, which are widely-used in binary neural networks literature~\cite{qin2020forward,lin2020rotated, xu2021learning}. CIFAR-10 is a common datasets for image classification, which contains 50k training images and 10k testing images with 10 different categories. Each image is of size 32x32 with RGB color channels. ImageNet ILSVRC-2012 is a large-scale dataset with over 120k training images and 50k verification images. Each image contains 224x224 resolutions with RGB color channels. It has 1000 different categories.

\begin{figure*}[h]
\centering
\includegraphics[width=0.80\linewidth]{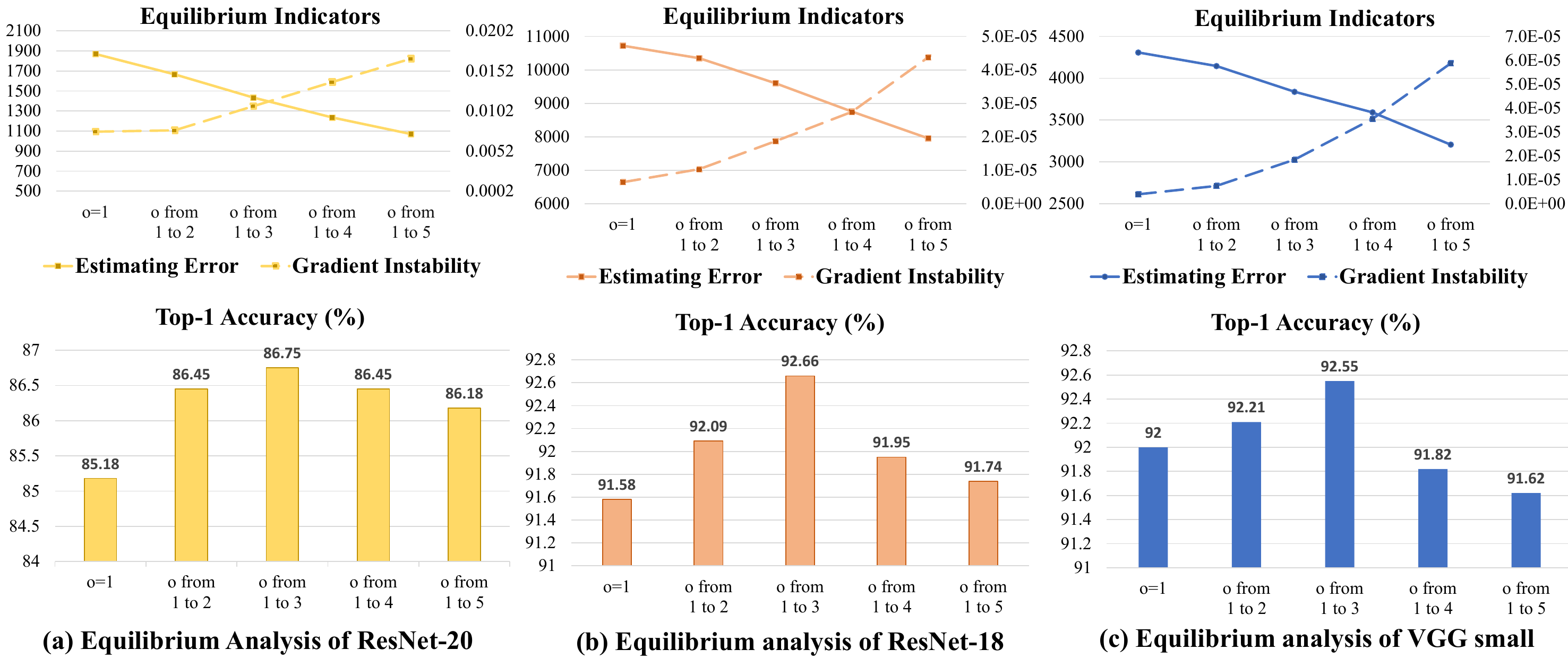}
\caption{Illustrations of the estimating error indicators (above), gradient instability indicators (above) and the Top-1 accuracy (below) at different scales of $o_\text{end}$ on CIFAR-10 dataset.}
\label{fig:equilibrium_analysis}
\end{figure*}

\textbf{Implementation Details.}  we follow the same setting as other binary methods~\cite{qin2020forward, lin2020rotated} used for fair comparison. For specific, we apply RandomCrop, RandomHorizontalFlip and Normalize for both CIFAR-10 and ImageNet pre-processing. We use SGD and set learning rate beginning from 0.1. Cosine learning rate descent schedule is adopted when training. What's more, we only use cross entropy as the loss function for classification. As for the hyper-parameter $o_{\text{end}}$, we set $o_{\text{end}}=3$ in all the experiments. We find that this value is suitable and robust to balance the estimating error and the gradient stability. Regarding the hyper-parameters $t$ and $i$ for gradient truncation, we simply set $t=1.5$ and $i=0.1$. All the models are implemented with PyTorch~\cite{paszke2019pytorch} on NVIDIA RTX3090 GPUs or NVIDIA RTX A6000 GPUs. For more details about the experiments parameters, please refer to our published codes and the README file in GitHub.

\subsection{Performance Study}\label{sec:performance}
To prove the performance of our method, we conduct
performance study in comparison with other binary methods. Note that our method only modify the estimators in backward process without other auxiliaries, e.g., additional modules or losses. To highlight the superiority of our approach, we add a column to note the auxiliaries used in other methods in the result tables.

We first test the performance of ReSTE on CIFAR-10~\cite{krizhevsky2009learning} with the SOTA methods. In detail, we binarize three backbone models, ResNet-18, ResNet-20~\cite{he2016deep} and VGG-small~\cite{simonyan2014very}. We compare a list of SOTA methods to validate our performance, including LBA~\cite{hou2016loss}, RAD~\cite{ding2019regularizing}, DSQ~\cite{gong2019differentiable}, Xnor-Net~\cite{rastegari2016xnor}, DoReFa-Net~\cite{zhou2016dorefa}, LQ-Net~\cite{zhang2018lq}, IR-Net~\cite{qin2020forward}, LCR-BNN~\cite{shang2022lipschitz}, RBNN~\cite{lin2020rotated}, FDA~\cite{xu2021learning}. For ResNet-20, we both evaluate the performance of our method in the basic ResNet architecture and the Bi-Real architecture~\cite{liu2018bi}. Experiments results are exhibited in Table \ref{CIFAR-10_performance}. From the table we can find that our ReSTE shows excellent performance, outperforming all the SOTA methods both at the setting of 1W/1A and 1W/32A without any assistance, e.g., modules or losses. For example, with ResNet-20 as the backbone, ReSTE respectively obtains 0.25\% and 0.45\% enhancement over the SOTA method RBNN~\cite{lin2020rotated} in the basic ResNet architecture and in the Bi-Real architecture~\cite{liu2018bi}, at the setting of 1W/1A, even that RBNN additionally adds a rotation module into the training. As for the setting of 1W/32A, ReSTE has 0.12\% improvement over the SOTA method LQ-Net~\cite{zhang2018lq}, which additional uses a Lipschitz loss to improve the training.

Moreover, we employ ReSTE on ResNet-18, ResNet-34~\cite{he2016deep} and validate the performance on large-scale ImageNet ILSVRC-2012~\cite{deng2009ImageNet}. In this setting, we compare ReSTE with ABC-Net~\cite{lin2017towards}, BWN~\cite{rastegari2016xnor}, TWN~\cite{alemdar2017ternary},  SQ-BWN and SQ-TWN~\cite{dong2017learning}, Xnor-Net~\cite{rastegari2016xnor}, HWGO~\cite{cai2017deep}, BWHN~\cite{hu2018hashing},  BNN+~\cite{darabi2018bnn+}, DoReFa-Net~\cite{zhou2016dorefa}, Bi-Real~\cite{liu2018bi}, Xnor-Net++~\cite{bulat2019xnor},  IR-Net~\cite{qin2020forward}, LCR-BNN~\cite{shang2022lipschitz}, RBNN~\cite{lin2020rotated}, FDA~\cite{xu2021learning}. 
At the setting of 1W/1A,  we use the Bi-Real architecture as most previous methods~\cite{qin2020forward, lin2020rotated, xu2021learning, tian2019contrastive} do for fair comparison. The results are shown in Table \ref{ImageNet_performance}. Similar as the analysis on CIFAR-10 dataset, ReSET also displays excellent performance and outperforms all the SOTA methods without any assistance, e.g., modules or losses. For example, with ResNet-18 as backbone, ReSTE has 0.68\% over the SOTA method FDA~\cite{xu2021learning}, at the setting of 1W/1A, even that FDA~\cite{xu2021learning} adds a noise adaptation module to help the training. About the 1W/32A setting, ReSTE also has 0.50 improvement over the SOTA method LQ-Net~\cite{zhang2018lq}, which has an additional loss to assist the training. 

To sum up, we can conclude that ReSTE has excellent performance and outperforms the SOTA methods in both CIFAR-10 and large-scale ImageNet ILSVRC-2012 datasets. The reason is that our ReSTE is always rational, with less estimating error than STE, as well as that we obtain the desirable degree of the equilibrium by the ReSTE, which is capable of flexibly balancing the estimating error and the gradient stability. Moreover, ReSTE surpasses other binary methods without any assistance of additional modules or losses, showing the importance of fully considering the gradient stability and finding the suitable degree of equilibrium to BNNs training.

\subsection{Estimators Comparison}
To further evaluate the effectiveness of our approach, we compare ReSTE with other estimators in the same and fair setting without other auxiliaries, e.g., modules or additional losses.

Specifically, we use ResNet-20 as our backbone, comparing ReSTE with STE~\cite{courbariaux2016binarized}, DSQ~\cite{gong2019differentiable}, EDE~\cite{qin2020forward}, FDA~\cite{xu2021learning}, RBNN~\cite{lin2020rotated}  on CIFAR-10~\cite{krizhevsky2009learning} at the setting of 1W/1A. Note that FDA here doesn't contain the noise adaptation module~\cite{xu2021learning} and RBNN doesn't use the rotation procedure since we only use the sign function with scalar in forward process for fair comparison. Experiments results are shown in Table \ref{ablation_table}. From the table we can observe that although ReSTE is concise, it significantly surpasses all the estimators in SOTA binary methods at the fully fair setting, with about 0.88\% and 0.95\% improvement over the estimators in RBNN and FDA. There are two facets of reasons. First is that our ReSTE always guarantees the rational property, with less estimating error than STE. Second is that we find out the desirable degree of the equilibrium with the assistance of the excellent ReSTE, which is capable of flexibly balancing the estimating error and the gradient stability. 

\begin{figure}[t]
\centering
\includegraphics[width=0.78\linewidth]{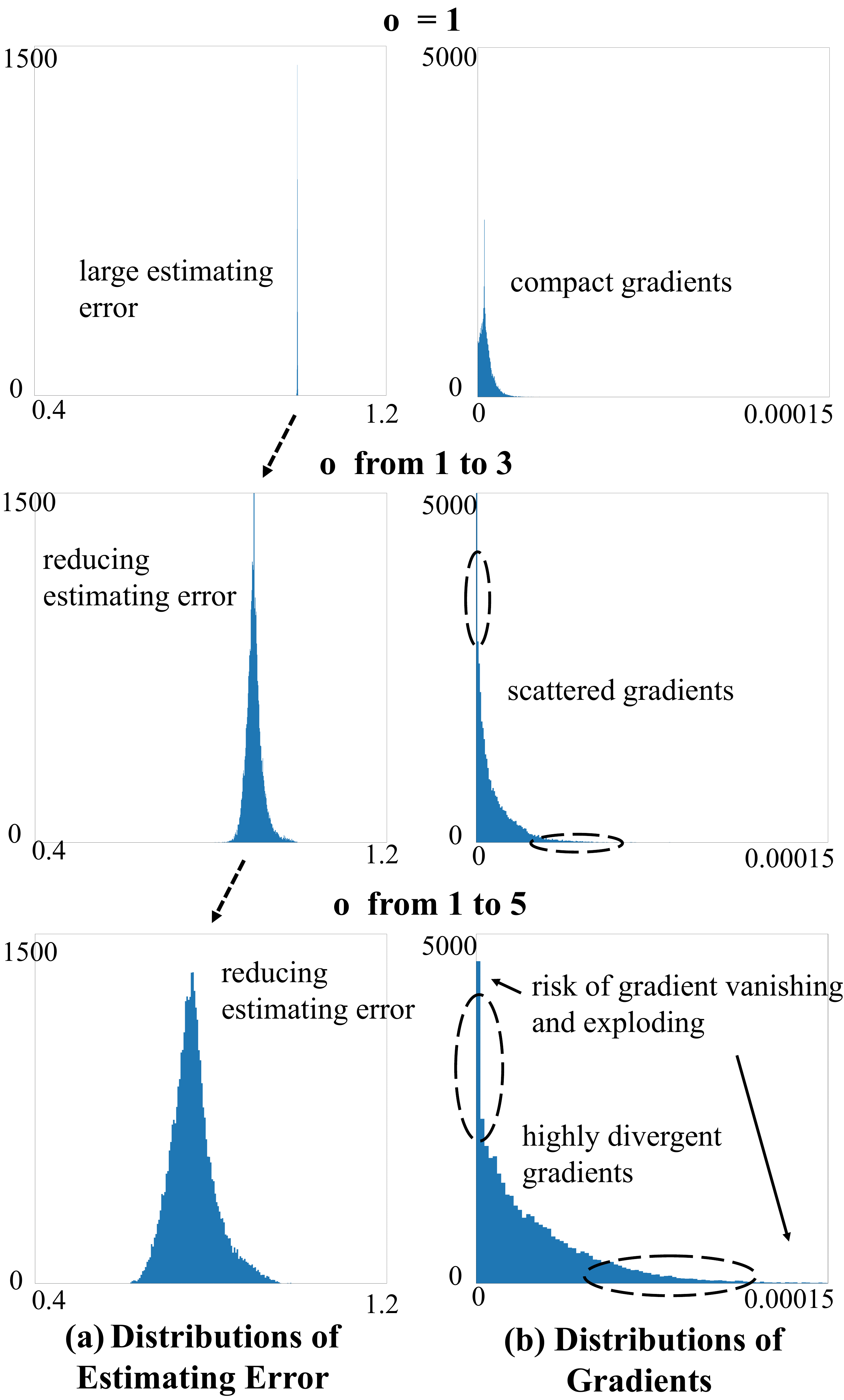}
\caption{Illustrations of distributions of the estimating error (left) and the gradients (right) at different scales of $o_\text{end}$. X-axes represent the values of the estimating error and the gradients, y-axes are the frequency.}
\label{fig:distributions}
\end{figure}

\begin{figure}[t]
\centering
\includegraphics[width=0.75\linewidth]{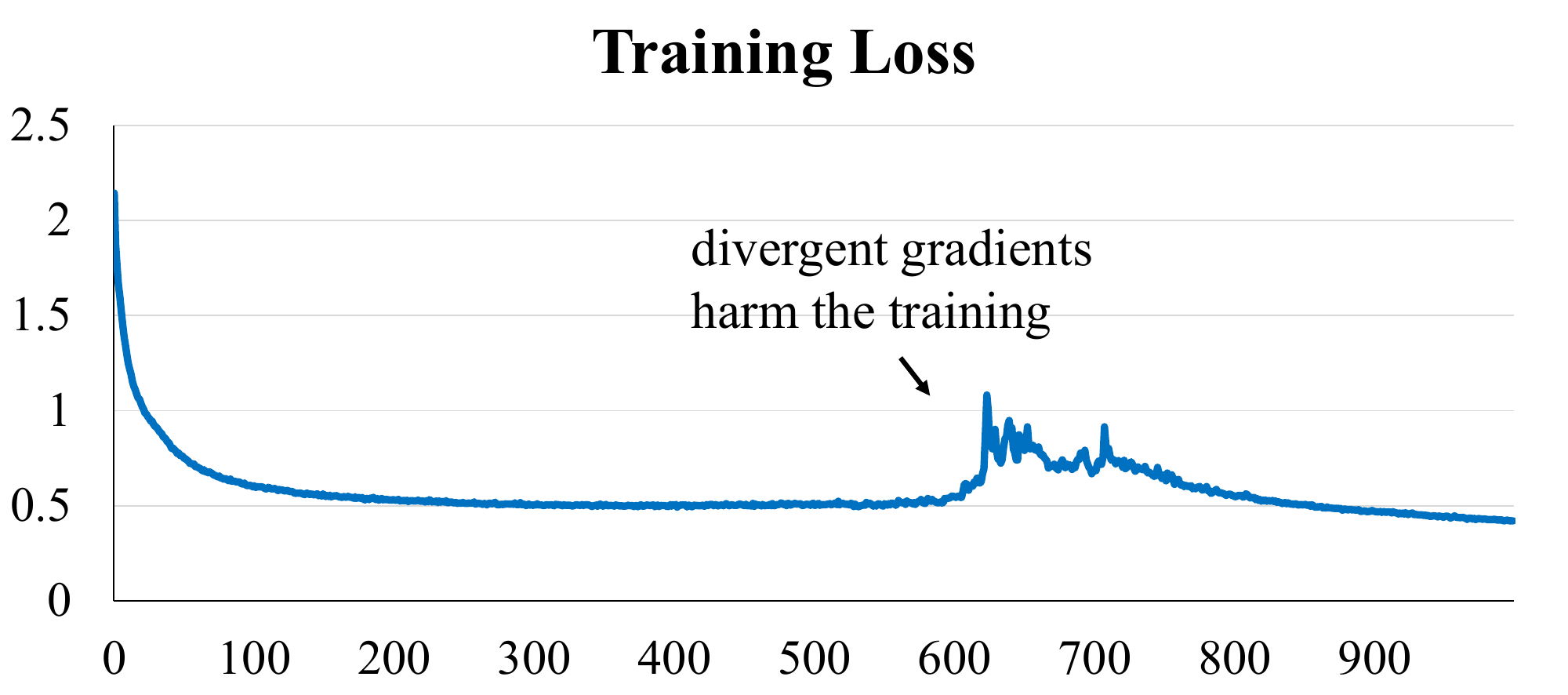}
\caption{Illustrations of an example that divergent gradients ($o_\text{end}=10$) will harm the BNNs training.}
\label{fig:training_fail}
\end{figure}

\subsection{Analysis of the Equilibrium Perspective}\label{sec:equilibrium_analysis}
To quantitatively and clearly demonstrate the equilibrium phenomenon and show the balancing ability of ReSTE, we adjust $o_{\text{end}}$ at different scales and meanwhile test the estimating error, gradient stability and the model performance. To make the results more convincing, we conduct the experiments with three widely-used backbones, ResNet-20, ResNet-18 and VGG-small. All the experiments are conducted on CIFAR-10 dataset at the setting of 1W/1A.
We evaluate the estimating error and gradient stability layers by layers with the indicators proposed in Sec. \ref{sec:equilibrium} and use the average results of all the binarized layers. Meanwhile, we will collect the results from different training epochs to obtain the final indicators for an overall training, as shown in Fig. \ref{fig:equilibrium_analysis}.

From the figures we can observe that with $o_{\text{end}}$ increasing, the estimating error becomes smaller and smaller, while the gradient instability becomes bigger and bigger. This observation shows that although the estimating error can be reduced by adjusting the estimator close to the sign function, the gradient stability will decline along with. In addition, the model performance increases first and then decreases with the change of $o_{\text{end}}$, which implies that the large gradient instability will harm the model performance. Such changes clearly reflect the equilibrium phenomenon and validate our claim that highly divergent gradients will harm the BNNs training.

In addition, it can also be seen from the figures that ReSTE can adjust the degree of equilibrium by easily changing the hyper-parameter $o_{\text{end}}$. Moreover, the desirable degrees of equilibrium, i.e., the desirable $o_{\text{end}}$ to produce high performance, are same in all the backbones, showing the robustness and universality of ReSTE. When applying ReSTE at different backbones for different applications, we can simply adjust $o_{\text{end}}$ to find out the suitable degree of the equilibrium and obtain good performance. More experiments about equilibrium analysis are shown in supplementary materials.

To obtain intuitive visualizations of the equilibrium phenomenon, we additional visualize the distributions of the estimating error and the distributions of gradient at different scales of $o$.  We use ResNet-18 as backbone and conduct the experiment on CIFAR10 dataset at the setting of 1W/1A. The results are shown in Fig. \ref{fig:distributions}. From the figure we can observe that with $o_{\text{end}}$ increasing, the peak values of the estimating error distribution become smaller, but the gradients become more divergent, which harms the model training and increases the risk of gradient vanishing or exploding. This visualization further demonstrate the equilibrium phenomenon and highlight the importance of finding the suitable degree of it.

To further validate our claim that highly divergent gradients will harm the model training, we demonstrate an example in Fig. \ref{fig:training_fail}. In this example, we use $o_\text{end}=10$ with ResNet-20 as backbone and test on CIFAR-10 dataset at the setting of 1W/1A. We can observe that the training loss has huge fluctuations at about 600 to 700 epochs due to the divergent gradients, causing the final accuracy decreases from 86.75 to 82.86. When $o_\text{end}$ further increase, the training will fail irreversible. This phenomenon verifies the harm of highly divergent gradients to model training and further demonstrates the importance of the equilibrium perspective.

\section{Conclusion}
\label{sec:conclusion}
In this work, we consider BNNs training as the equilibrium between the estimating error and the gradient stability. In this view, we firstly design two indicators to quantitatively and clearly demonstrate the equilibrium phenomenon. In addition, to balance the estimating error and the gradient stability well, we look back to the original STE and revise it into a new power function based estimator, rectified straight through estimator (ReSTE). Comparing to other estimators, ReSTE is rational and is capable of flexibly balancing the estimating error and the gradient stability. Extensive performance study on two datasets have demonstrated the effectiveness of ReSTE, surpassing state-of-the-art methods. By two carefully-designed indicators, we demonstrate the equilibrium phenomenon and shows the ability of ReSTE to adjust the degree of equilibrium. 

\section{Acknowledgments}
This work was supported partially by the NSFC (U21A20471, U1911401, U1811461), Guangdong NSF Project (No. 2023B1515040025, 2020B1515120085).

\clearpage

{\small
\bibliographystyle{ieee_fullname}
\bibliography{egbib}
}

\end{document}